\documentclass[conference]{IEEEtran}
\IEEEoverridecommandlockouts
\usepackage{cite}
\usepackage{comment}
\usepackage{amsmath,amssymb,amsfonts}
\usepackage{soul}
\usepackage{algorithmic}
\usepackage{graphicx}
\usepackage{textcomp}
\usepackage{multirow}
\usepackage{hyperref}
\usepackage[table]{xcolor}
\usepackage{xurl}
\usepackage[labelfont=bf,tableposition=top,skip=2pt]{caption}
\usepackage{xcolor}
\def\BibTeX{{\rm B\kern-.05em{\sc i\kern-.025em b}\kern-.08em
    T\kern-.1667em\lower.7ex\hbox{E}\kern-.125emX}}
    
\newcommand{\XD}[1]{}
\renewcommand{\XD}[1]{{\color{magenta} XD: {#1}}}
\newcommand{\ram}[1]{}
\renewcommand{\ram}[1]{{\color{blue} RamV: {#1}}}
\newcommand{\miracle}[1]{}
\renewcommand{\miracle}[1]{{\color{red} Miracle: {#1}}}
    
\begin{document}

\title{Leveraging Trajectory Prediction for Pedestrian Video Anomaly Detection
}

\author{\IEEEauthorblockN{Asiegbu Miracle Kanu-Asiegbu}
\IEEEauthorblockA{\textit{Mechanical Engineering} \\
\textit{University of Michigan}\\
Ann Arbor, MI, USA \\
akanu@umich.edu}
\and
\IEEEauthorblockN{Ram Vasudevan}
\IEEEauthorblockA{\textit{Mechanical Engineering} \\
\textit{University of Michigan}\\
Ann Arbor, MI, USA \\
ramv@umich.edu}
\and
\IEEEauthorblockN{Xiaoxiao Du}
\IEEEauthorblockA{\textit{Naval Architecture and Marine Engineering} \\
\textit{University of Michigan}\\
Ann Arbor, MI, USA \\
xiaodu@umich.edu}
}

\maketitle

\begin{abstract}
Video anomaly detection is a core problem in vision. Correctly detecting and identifying anomalous behaviors in pedestrians from video data will enable safety-critical applications such as surveillance, activity monitoring, and human-robot interaction. In this paper, we propose to leverage trajectory localization and prediction for unsupervised pedestrian anomaly event detection. Different than previous reconstruction-based approaches, our proposed framework rely on the prediction errors of normal and abnormal pedestrian trajectories to detect anomalies spatially and temporally. We present experimental results on real-world benchmark datasets on varying timescales and show that our proposed trajectory-predictor-based anomaly detection pipeline is effective and efficient at identifying anomalous activities of pedestrians in videos. Code will be made available at \url{https://github.com/akanuasiegbu/Leveraging-Trajectory-Prediction-for-Pedestrian-Video-Anomaly-Detection}. 
\end{abstract}


\section{Introduction}
Anomaly detection aims to detect, identify, and recognize events that do not conform to expected activities \cite{popoola2012video}. Understanding and detecting abnormal events from video data is critical for real applications such as security surveillance, video annotation and indexing, human activity recognition and monitoring, and safe interactions between humans/pedestrians and autonomous systems (e.g., mobile robots and autonomous vehicles). Anomaly patterns can be diverse, context-dependent, and usually in relatively smaller quantities than normal behaviors in collected data. Thus, the anomaly detection problem is typically formulated as learning a representation of normality from large amounts of training data and anomalies are determined if a test sample does not fit well with the learned ``normal'' representations. 

There exists a large body of literature for video anomaly detection. Traditional methods are often rule-based or rely on low-level/hand-crafted features. Amit et al. \cite{amit2008} used multiple local features at different spatial levels to detect anomalies for surveillance-type applications. However, it lacks the ability to monitor sequential actions over longer term and requires manual thresholding. Mahadevan et al. \cite{vijay2010} used a mixture of dynamic mixture (MDT, \cite{chan2008modeling})-based representation to extract appearance and dynamics features for anomaly detection in crowded scenes. However, the computation time is quite long for this method (testing time is about 25 seconds per frame). Various other techniques, such as Projection to Latent Structures (PLS) regression \cite{fouzi2015enhanced}, sparse coding \cite{bin2011,cewu2013}, optical flow \cite{tan2016fast},  histograms \cite{zaharescu2010anomalous}, and Markov Random Fields \cite{ide2016sparse}, have also been used for anomaly detection. 

Deep learning (DL) methods have gained popularity and shown satisfactory performance in recent years for anomaly detection applications. The most common DL approaches are based on the concept of reconstruction, where the network learns to reconstruct the frames and/or motion features present in the data. The assumption is that the network will be able to reconstruct regular/normal dynamics in training data with low reconstruction errors but will not accurately reconstruct irregular frames/motions. Techniques such as autoencoders \cite{hasan2016learning}, spatiotemporal autoencoders \cite{chong2017abnormal}, convolutional long short-term memory (Conv-LSTM) \cite{weixin2017}, generative adversarial network (GAN) \cite{ravanbakhsh2017abnormal}, U-Net with memory module \cite{park2020learning}, and convolutional neural network (CNN)-based data augmentation \cite{li2021cutpaste} have been 
developed to reconstruct normal data. Anomalies are indicated with a high data reconstruction error.

In this paper, instead of relying on reconstruction error, we propose to identify anomalous events by evaluating their expectation and introduce a prediction-based pipeline for pedestrian video anomaly detection. Liu et al. \cite{liu2018future} was the first work, to our knowledge, that leverages the difference between a predicted future frame and its ground truth to detect an abnormal event. The authors stated that since anomalies are defined as events that do not conform to the expectation, it is more natural and intuitive to use prediction-based approaches for identifying anomalies. Our work differs from \cite{liu2018future} in that instead of predicting the entire video frame, we propose to take advantage of recent advances in object/human localization and trajectory predictors and use predicted human trajectories to detect anomalous events. Morais et al. \cite{Morais_2019_CVPR} proposed a recurrent neural network that leverages 2D human skeleton trajectories for detecting abnormal events. Our work is similar but will use a conditional variational autoencoder (CVAE) \cite{sohn2015learning} based novel architecture for trajectory prediction and will use a simpler feature representation (only bounding box coordinates of pedestrian trajectories, instead of full-body joints). Ionescu et al. \cite{ionescu2019object} developed an object/human-centric approach based on the convolutional auto-encoders that learns appearance features of the pedestrians in the scene and proposed to train a one-vs-rest support vector machine to classify the abnormal and normal test samples. Our work also consists of a human localization and detection step, but will incorporate three entirely unsupervised error measures to detect anomalies and will not require frame-level or human-level anomaly labels during training. 

This paper is organized as follows. Section I introduces the problem of video anomaly detection,  reviews related literature and motivates our proposed work. Section II describes our proposed pipeline with human localization, trajectory prediction, and anomaly detection steps. Section III presents our anomaly detection results on two real-world benchmark datasets and provide ablation study and analysis on effectiveness of the trajectory predictor for anomaly detection on varying trajectory sequence lengths (timescales).
Section IV presents our conclusions and discussions on future work.





\section{Proposed Framework}
\begin{figure*}[t!]
  \centering
  \includegraphics[width=\textwidth,trim=0 0 0 0,clip]{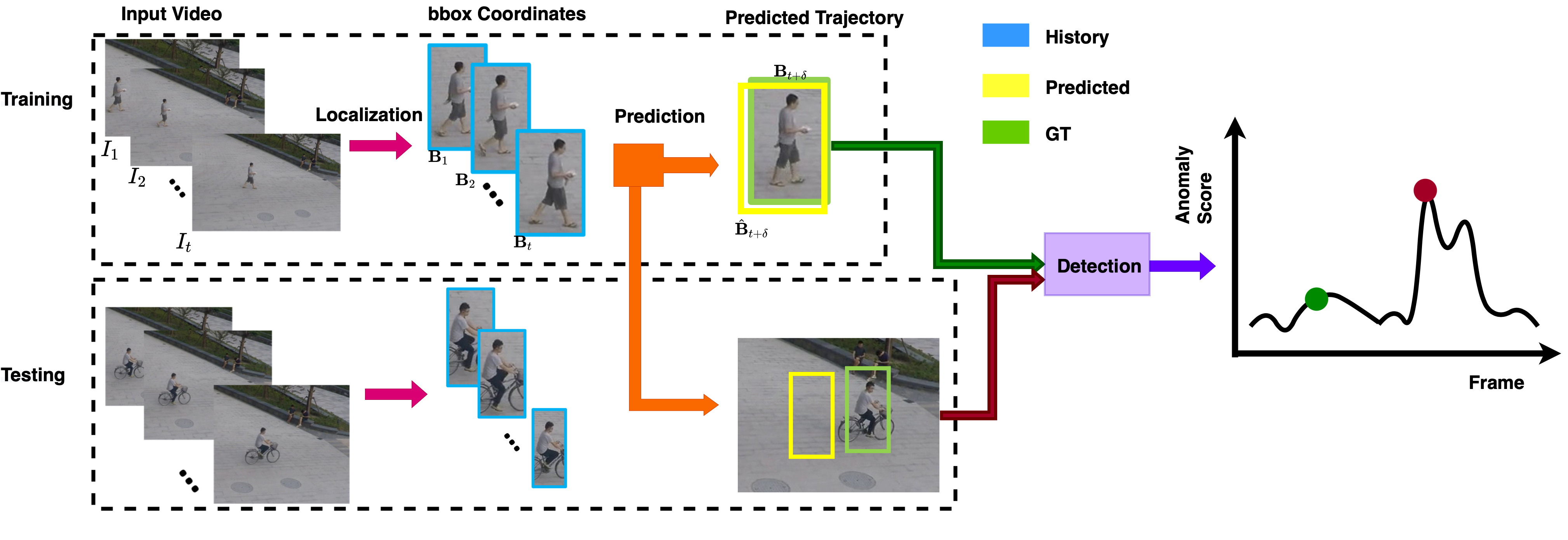}  
  \caption{Overview of our proposed trajectory prediction based video anomaly detection pipeline. In the localization step, Deep SORT is used to detect and track humans (pedestrians) in the scene (magenta arrow). In the prediction step, BiTraP learns normal activities from training data and performs inference on testing data to generate future trajectory predictions (orange block). In the detection step, the predicted trajectories are compared with the ground truth bounding boxes to compute the anomaly score (purple block). The input of the pipeline is training and testing video clips and the output of the pipeline is the anomaly score at each frame and the spatial location of anomalous activities.}
  \label{fig:methodoverview}
\end{figure*}

In this section, we describe our proposed trajectory prediction based anomaly detection framework in detail. Figure~\ref{fig:methodoverview} shows an overview of the pipeline. The anomaly detection problem is set up with a training set and a testing set, where the videos in training only contains ``normal'' events and no anomalies. Testing videos can contain both normal and abnormal/anomalous activities and our goal is detect the anomalies in the testing set. The input data in our work are in the form of 2-D RGB camera videos containing human/pedestrian activities, such as walking, running, jumping, etc. The output of our method is the detected anomaly scores for each frame in the video. Our pipeline consists of three stages, spatial localization, future trajectory prediction, and anomaly detection. Section~\ref{sec:inputdeepsort} describes how we extract and localize human trajectory information from input videos, Section~\ref{sec:predictor} describes our trajectory prediction method, and Section~\ref{sec:detector} describes how we use trajectory prediction to  perform anomaly detection.


\subsection{Input Trajectory Localization}
\label{sec:inputdeepsort}
Given an input video containing human activities, an object (human) tracking algorithm can be used to detect, localize and track the spatial location of each pedestrian in the scene. In this work, the Deep SORT \cite{wojke2017simple} approach is used to extract pedestrian trajectories for its simplicity and effective detection and tracking performance in real-time. Deep SORT detects the human bounding box locations, aspect ratio, and velocities at each frame and the Kalman filter is used to form a tracking sequence from the detections \cite{bewley2016simple}. Deep SORT also takes advantage of the appearance and motion in the images by using a convolutional neural network to extract feature vectors from all the image crops of the detected human bounding boxes. The outputs of Deep SORT are the tracked sequences of each pedestrian in bounding box coordinates (top left and bottom right). We then convert the coordinate representation to a $4\times1$ vector of $\mathbf{B}_t^i=[x_t^i, y_t^i, w_t^i, h_t^i]$, where $x_t^i$ and $y_t^i$ are the coordinate values at the center of the bounding box and $w_t^i, h_t^i$ are the width and height of the bounding box for pedestrian $i$ at frame $t$. Then, the localized pedestrian trajectory sequences will be used as input for future trajectory prediction and the predictions will be used for anomaly event detection of pedestrians in the video.

\subsection{Trajectory Prediction: BiTraP}
\label{sec:predictor}
We propose BiTraP \cite{yao2021bitrap}, a goal-conditioned \textbf{Bi}-directional \textbf{tra}jectory \textbf{p}rediction algorithm based on the conditional variational autoencoder (CVAE) \cite{sohn2015learning}, for the trajectory prediction module in our pipeline.  Similar to a standard CVAE, the BiTraP method contains an encoder and a decoder structure. Different from prior work such as~\cite{lee2017desire,bhattacharyya2018accurate,mangalam2020not}, the BiTraP algorithm extends the standard recurrent neural network encoder-decoder based CVAE trajectory predictor by incorporating goal (endpoints of trajectory) estimation to improve trajectory prediction performance. 

\begin{figure*}[h!]
  \centering
  \includegraphics[width=0.9\textwidth,trim=0 0 0 0,clip]{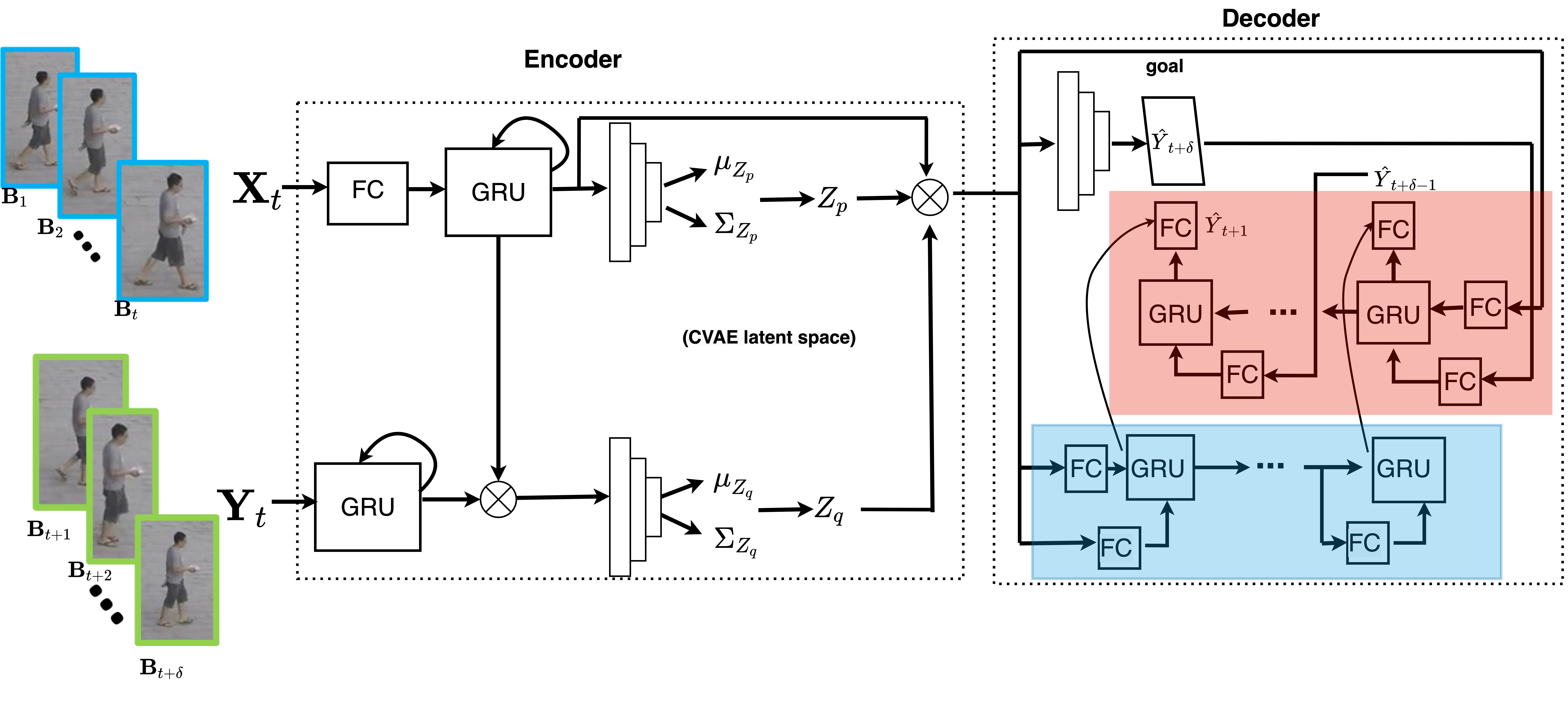}  
  \caption{BiTraP network architecture. In the encoder, a set of fully connected (FC) layer and gated recurrent units (GRUs) extract features from the observed $\mathbf{X}_t$ and target $\mathbf{Y}_t$ trajectories, and a three-layer multi-layer perceptron (MLP) is used to learn the parameters $\mu$ and $\Sigma$ for the Gaussian latent variable $Z$. In the decoder, another three-layer MLP is used to generate future goals (the endpoints of the trajectories) $\hat{\mathbf{Y}}_{t+\delta}$. The light blue shaded region shows a forward RNN with stacked FCs and GRUs, which makes predictions at each step from current position to estimated goals. The red shaded region shows a backward RNN, which propagates from the goals to current position. Both forward and backward RNNs form the novel bi-directional decoder used in BiTraP. A more detailed illustration can also be seen in \cite{yao2021bitrap}.}
  \label{fig:diagram2} 
\end{figure*}

Let $\mathbf{X}_t^i=[\mathbf{B}_{t-\tau+1}^i,\mathbf{B}_{t-\tau+2}^i,...,\mathbf{B}_{t}^i]$ denote observed past trajectory of pedestrian $i$ at time $t$, let $\mathbf{Y}_t^i=[\mathbf{B}_{t+1}^i,\mathbf{B}_{t+2}^i,...,\mathbf{B}_{t+\delta}^i]$ denote the ground truth future trajectory of pedestrian $i$, and let $\hat{\mathbf{Y}}_t^i=[\hat{\mathbf{B}}_{t+1}^i,\hat{\mathbf{B}}_{t+2}^i,...,\hat{\mathbf{B}}_{t+\delta}^i]$ denote the predicted future trajectory of pedestrian $i$, where $\tau$ and 
$\delta$ are observation and prediction horizons, respectively.  The $\mathbf{X}_t^i$ and $\mathbf{Y}_t^i$ can be obtained from Section~\ref{sec:inputdeepsort}, and BiTraP is used to obtain predicted $\hat{\mathbf{Y}}_t^i$ for all pedestrians given $\mathbf{X}_t^i$ and $\mathbf{Y}_t^i$, $\forall i$. Figure~\ref{fig:diagram2} shows the network architecture of BiTraP. Given observed bounding box trajectories $\mathbf{X}_t$,  a gated-recurrent unit (GRU) encoder network is first used to obtain encoded feature vector $h_t$. In training, the ground truth future trajectories $\mathbf{Y}_t$ are encoded by another GRU, yielding feature vector $h_{Y_t}$. We assume the latent variable follows a Gaussian distribution, $Z\sim\mathcal{N}(\mu_{Z}, \Sigma_{Z})$, where the parameters of $\mu_{Z_p}, \Sigma_{Z_p}$ and $\mu_{Z_q}, \Sigma_{Z_q}$ can be learned from the feature vectors  $h_t$ and $h_{Y_t}$. The latent variable $Z_p$ represents information from the historic (observed) trajectory sequences, and the variable $Z_q$ captures dependencies between the past (observed) and ground truth future trajectories. 

BiTraP contains a novel bi-directional decoder, which consists of a forward and a backward recurrent neural network (RNN) to propagate from current position to estimated future goals (endpoints of trajectories) and from the goals to current position. In the decoder, the latent variable and the feature vectors are concatenated and passed through a three-layer multi-layer perceptron to generate future predictions of goals $\hat{\mathbf{Y}}_{t+\delta}$. The forward RNN then predicts the trajectory locations at each future timestep (waypoints) by using stacked fully connected layer and GRUs, whereas the backward RNN takes the predicted goal $\hat{Y}_{t+\delta}$ as the initial input and updates the hidden states of the backward GRUs from the goal (time $t+\delta$) to the current location ($t+1$). Eventually, the hidden states from both forward and backward RNNs are concatenated at each time step to generate the final predicted trajectory locations at each frame $t\in\{t+1, ..., t+\delta\}$.

Recall that in our video anomaly detection problem, training videos contain only normal activities (such as pedestrian walking) and testing videos can contain both normal and abnormal/anomalous activities (such as jumping or biking). In the trajectory prediction process, BiTraP is first trained on training video clips to learn the trajectory pattern and motion of trajectories in normal activities. Then, the trained BiTraP model is used to perform inference on the testing trajectories extracted from testing videos. Our assumption, per the definition of ``anomaly'' as described in Section I, is that anomalous events differ from expectations. Thus, our hypothesis is that the BiTraP model, trained on normal activities in the training data, will be able to characterize the distribution of normal walking scenarios and predict well on normal walking activities, whereas the predicted trajectory locations will differ greatly from the ground truth when testing on anomalous activities. Simply put, we assume that normal events can be predicted well and anomalies cannot.  We will later provide experimental results to demonstrate the performance of our predictor, which verified this hypothesis.

\subsection{Anomaly Event Detection}
\label{sec:detector}
As discussed above, our hypothesis is that the predicted trajectories of pedestrians performing anomalous activities will deviate further from the ground truth trajectory compared to normal activities.  Thus, the trajectory prediction error between the predicted trajectory bounding box and the ground truth bounding box locations can be used as an indicator for identifying anomalous events. 

\subsubsection{Prediction Error Measures}
In this work, we propose to use three types of unsupervised error measures to compute the trajectory prediction errors: the intersection over union (IOU) \cite{lin2014microsoft}, the generalized intersection over union (GIOU) \cite{rezatofighi2019generalized}, and L2-norm. The IOU is the most popular evaluation metric for object detection \cite{rezatofighi2019generalized} and computes the overlap between predicted and ground truth bounding boxes. The GIOU extends upon IOU to cases where the prediction and ground truth bounding boxes may be non-overlapping, and is a good alternative error measure. Both IOU and GIOU are invariant to the scale of the bounding boxes, and higher IOU and GIOU values indicate there are more overlap between the predicted and ground truth bounding boxes (i.e., the predictions match the ground truth more accurately). The L2-norm errors are computed by taking the Euclidean distance between the predicted and ground truth bounding box coordinates, and smaller L2 error indicates a better prediction. Thus, based on our assumption above that normal events can be well predicted and anomalies cannot, we use $1-IOU$, $1-GIOU$, and L2 error to detect the anomaly. Mathematically, the three measures can be written as
\begin{equation}
    m_1(t)^i = 1- IOU = 1- \frac{|\mathbf{B}_t^i\cap \hat{\mathbf{B}}_t^i|}{|\mathbf{B}_t^i\cup \hat{\mathbf{B}}_t^i|};
\label{eq:m1}
\end{equation}
\begin{align}
\begin{split}
     &m_2(t)^i = 1- GIOU = 1- \left [ \frac{|\mathbf{B}_t^i\cap \hat{\mathbf{B}}_t^i|}{|\mathbf{B}_t^i\cup \hat{\mathbf{B}}_t^i|} - \frac{|C_t^i\backslash (\mathbf{B}_t^i\cup \hat{\mathbf{B}}_t^i)|}{|C_t^i|} \right ], \\
    &\mbox{\small{where $C_t^i$ is the smallest enclosing  bounding box for $\mathbf{B}_t^i$ and $\hat{\mathbf{B}}_t^i$;}}
\end{split}
\label{eq:m2}
\end{align}
\begin{equation}
    m_3(t)^i = \left \| \mathbf{B}_t^i -  \hat{\mathbf{B}}_t^i \right \|_2,
\label{eq:m3}
\end{equation}
where $\mathbf{B}_t^i$ and $\hat{\mathbf{B}}_t^i$ are the ground truth (as localized from Section~\ref{sec:inputdeepsort}) and the predicted (as predicted from Section~\ref{sec:predictor}) bounding boxes at time $t$ for the $i^{th}$ pedestrian. As described above, higher $m_1$, $m_2$, $m_3$ values correspond to increased probability of the occurrence of an anomalous event.  

\subsubsection{Sub-sequence Handling}
\label{subsec:sub-sequence}
Our BiTraP predictor handles trajectory sequences as inputs and outputs (i.e., the observed and prediction horizons $\tau$ and $\delta \geq 1$). In our experiments, we perform trajectory prediction using a sliding window of one. That is to say, if a training pedestrian video sequence has a total of N frames, BiTraP first takes $[\mathbf{B}_{1}^i,\mathbf{B}_{2}^i,...,\mathbf{B}_{\tau}^i]$ as input to predict $[\hat{\mathbf{B}}_{\tau+1}^i,\hat{\mathbf{B}}_{\tau+2}^i,...,\hat{\mathbf{B}}_{\tau+\delta}^i]$ as outputs. Then, the sequence $[\mathbf{B}_{2}^i,\mathbf{B}_{3}^i,...,\mathbf{B}_{\tau+1}^i]$ is used as input to predict $[\hat{\mathbf{B}}_{\tau+2}^i,\hat{\mathbf{B}}_{\tau+3}^i,...,\hat{\mathbf{B}}_{\tau+1+\delta}^i]$. This process continues until the total N frames has been exhausted, i.e., when  $[\hat{\mathbf{B}}_{N-\delta+1}^i,\hat{\mathbf{B}}_{N-\delta+2}^i,...,\hat{\mathbf{B}}_{N}^i]$ is predicted given $[\mathbf{B}_{N-\delta-\tau+1}^i,\mathbf{B}_{N-\delta-\tau+2}^i,...,\mathbf{B}_{N-\delta}^i]$. Then, for each of the predicted trajectory sub-sequences, error measure values $m$ can be computed against the ground truth following \eqref{eq:m1}-\eqref{eq:m3}.

\begin{figure}[h!]
  \centering
  \includegraphics[width=\columnwidth,trim=0 2.5cm 0 0cm,clip]{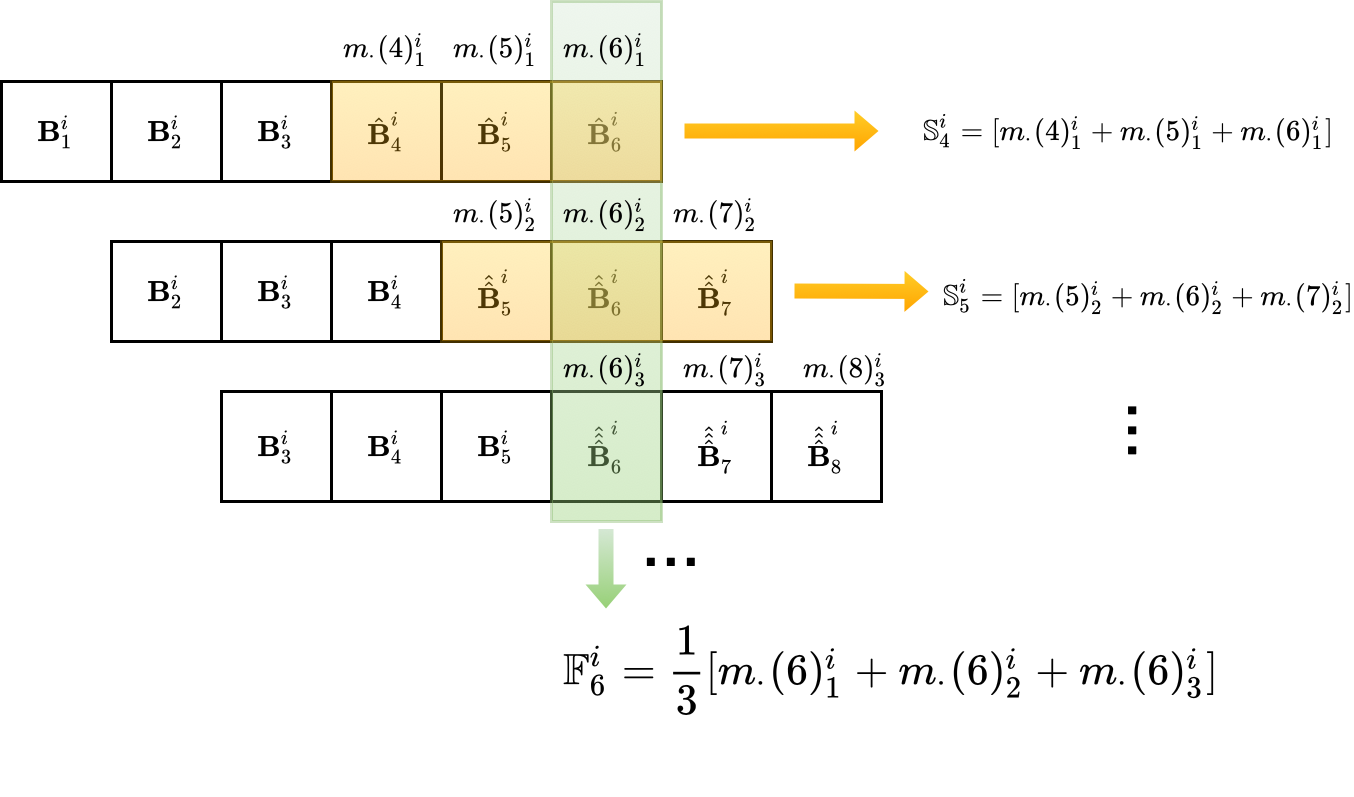}
  \caption{An example illustrating the two methods of aggregating error measures for sub-sequences, the ``summed error'' $\mathbb{S}$ and ``flattened error'' $\mathbb{F}$. In this example, $N=8,\tau=\delta=3$. The variable $\mathbf{B}$ refers to the observed trajectory sequences and $\hat{\mathbf{B}}$,$\hat{\hat{\mathbf{B}}}$, and $\hat{\hat{\hat{\mathbf{B}}}}$ refer to the predictions given each sliding window of observed sequences. The summed error sums up the error measurement $m_\cdot$ for all predictions, and the flattened error averages across the multiple predictions at the same timestep. }
  \label{fig:diagram3} 
\end{figure}

We will explain further using a simple example. Figure~\ref{fig:diagram3} shows a visual illustration of the example. Assume we have a pedestrian sequence of $N=8$ frames in total (pedestrian index is $i$). Assume observed sequence length is $\tau=3$ and we are predicting the next $\delta=3$ frames. Then, the 8-frame-length sequence can be broken down into three sub-sequences, if the sliding window size is 1: $[\mathbf{B}_{1}^i,\mathbf{B}_{2}^i,\mathbf{B}_{3}^i] \rightarrow
[\hat{\mathbf{B}}_{4}^i,\hat{\mathbf{B}}_{5}^i,\hat{\mathbf{B}}_{6}^i]$; $[\mathbf{B}_{2}^i,\mathbf{B}_{3}^i,\mathbf{B}_{4}^i] \rightarrow
[\hat{\hat{\mathbf{B}}}_{5}^i,\hat{\hat{\mathbf{B}}}_{6}^i,\hat{\hat{\mathbf{B}}}_{7}^i]$; and $[\mathbf{B}_{3}^i,\mathbf{B}_{4}^i,\mathbf{B}_{5}^i] \rightarrow
[\hat{\hat{\hat{\mathbf{B}}}}_{6}^i,\hat{\hat{\hat{\mathbf{B}}}}_{7}^i,\hat{\hat{\hat{\mathbf{B}}}}_{8}^i]$, where $\rightarrow$ means to predict. Then, the error measure between the predictions and the ground truth can be computed for each of these predictions according to \eqref{eq:m1}-\eqref{eq:m3}, and we obtain three sets of error measures: $[m_\cdot(4)^i_1, m_\cdot(5)^i_1,m_\cdot(6)^i_1]$, $[m_\cdot(5)^i_2, m_\cdot(6)^i_2,m_\cdot(7)^i_2]$, and $[m_\cdot(6)^i_3, m_\cdot(7)^i_3,m_\cdot(8)^i_3]$, where $m_\cdot \in \{m_1, m_2, m_3\}$ as in \eqref{eq:m1}-\eqref{eq:m3}. The subscript $p\in\{1,2,3\}$ for $m_\cdot(t)^i_p$ refers to each of the three sub-sequences (since $\hat{\mathbf{B}}_{6}^i$ may have different prediction value than $\hat{\hat{\mathbf{B}}}_{6}^i$, for example, $m_\cdot(6)^i_1$ may be different than $m_\cdot(6)^i_2$ as well, thus the subscript indices are used to clarify this difference for each sub-sequence). 

As seen, using a sliding window can yield multiple predictions at certain timesteps (e.g., three predictions at timestep $t=6$ in this example). To address this issue, we developed two ways to combine and evaluate the errors, named the ``summed error'' $\mathbb{S}$ and ``flattened error'' $\mathbb{F}$, where the summed error sums up prediction errors for all prediction timesteps in each sub-sequence, and the flattened error averages prediction errors at the same timestep across sub-sequences.  Mathematically, the two types of errors can be written as $\mathbb{S}_{t_c}^i = \sum_{t=t_c}^{t_c+\delta} m_\cdot(t)^i_p$ and $\mathbb{F}_t^i = \frac{1}{P} \sum_{p=1}^{P} m_\cdot(t)^i_p$, where $t_c$ is the first timestep to predict and $P$ is the total number of sub-sequences with the sliding window. This way, we have both a ``horizontal'' and ``vertical'' way of aggregating the multiple predictions and the $\mathbb{S}_{t}^i$ and $\mathbb{F}_{t}^i$ are regarded as the anomaly score for each pedestrian $i$ at each frame (timestep) $t$.


\subsubsection{Frame-level Anomaly Score Computation} After the process above, we know the predicted bounding box locations for each pedestrians and obtain $\mathbb{S}_{t}^i$ and $\mathbb{F}_{t}^i$ (anomaly scores) for each pedestrian $i$ at each frame (timestep) $t$. However, there may be multiple pedestrians in a frame. Thus, to compare with the ground truth frame-level anomaly labels provided in the datasets and to provide frame-level anomaly event detection evaluations comparable to literature such as \cite{hasan2016learning,liu2018future, Morais_2019_CVPR}, we need to convert the pedestrian-level errors to frame-level anomaly scores. Following \cite{Morais_2019_CVPR}, we use a max pooling operator on each frame and use the highest anomaly score across all pedestrian instances to represent the frame-level anomaly score. At each frame, the anomaly score is computed as
\begin{equation}
    \mathcal{S}_t = \max_{i} \mathbb{S}_{t}^i, \mathcal{F}_t = \max_{i} \mathbb{F}_{t}^i.
\end{equation}
This way, the highest anomaly score will be used to represent the frame and the influence of normal trajectories can be suppressed, especially when the number of normal versus anomalous events vary for each scene \cite{Morais_2019_CVPR}. The frame-level anomaly scores $\mathcal{S}_t$ and $\mathcal{F}_t$ for all $t$ are the final outputs of our pipeline for anomaly detection.

\subsection{Implementation Details}
For the localization step, we follow the Deep SORT with YOLO v4 \cite{bochkovskiy2020yolov4} implementation as in {\url{https://github.com/LeonLok/Deep-SORT-YOLOv4/tree/master/tensorflow2.0/deep-sort-yolov4}}. For the BiTraP predictor, a hidden size of 256 for both encoders and decoders were used, the learning rate is 0.001 and changes with plateau scheduler, as suggested in \cite{yao2021bitrap}. The number of predicted future trajectory is one ($K=1$), and the batch size is 30.

\section{Experimental Results}
In this section, we present quantitative and qualitative experimental results and discussions on computation time.

\subsection{Datasets and Experimental Setup}
We evaluate our proposed pipeline on two real-world benchmark anomaly event detection datasets, the \textit{Avenue} dataset \cite{cewu2013} and the \textit{ShanghaiTech} dataset \cite{luo2017revisit}. The \textit{Avenue} dataset was collected in CUHK campus avenue with 16 training and 21 testing video clips (15,328 training frames and 15,324 testing frames in total at 25 frames per second, fixed camera angle). Each frame in \textit{Avenue} is $640\times360$ pixels. The anomalous events seen in \textit{Avenue} includes running and throwing objects. The ground truth anomaly event labels are provided spatially and temporally in the \textit{Avenue} dataset. The \textit{ShanghaiTech} dataset captures 13 scenes on ShanghaiTech university campus with complex light conditions and varying camera angles. It contains 274,515 training frames and 42,883 testing frames with 130 abnormal events, such as chasing, brawling, loitering and riding a bike. Each frame in \textit{ShanghaiTech} is $856\times480$ pixels. The ground truth of the spatial and temporal location of the abnormal events in the frames is also annotated and provided in the \textit{ShanghaiTech} dataset. 

Each dataset contains a training set and testing set of video clips, where only ``normal'' activities were included in the training videos and both normal and anomalous events were captured in the testing videos. Thirty percent of the training set was set aside for validation. In our experiments, Deep SORT was first used to localize and extract bounding box information for each pedestrian in both training and testing videos. Then, the BiTraP predictor was trained given the trajectory sequences in the training dataset only in order to learn the patterns of normal activities.  
In testing, the trained BiTraP models were used to predict future trajectories for pedestrians in the testing dataset and frame-level anomaly scores were obtained by computing the error measures between the ``ground truth'' (as extracted by Deep SORT) trajectories and the predicted trajectories.  We use frame-level ROC (receiver operating characteristic) curve and the AUC (area under curve) as benchmark evaluation metrics following \cite{hasan2016learning, Morais_2019_CVPR} to compare and report the anomaly detection results. 

\begin{table*}[t!]
\caption{Frame-level AUC for Avenue and ShanghaiTech datasets using our proposed pipeline with BiTraP versus stacked LSTM as trajectory predictor. For each dataset, the detection results using the summed error $\mathcal{S}$ and the flattened error $\mathcal{F}$ are reported with three measures 1-IOU ($m_1$), 1-GIOU ($m_2$) and L2 norm ($m_3$) are reported. The \textcolor{blue}{blue} highlights the best performance in LSTM method and the \textcolor{red}{red} highlights the best performance in BiTraP method for each of the summed error and the flattened error measures. The \textbf{bold} highlights the best performance overall.} 
\label{table:frame_auc_error_all} 
\centering
\begin{tabular}{c|c|ccc|ccc|ccc|ccc} 
 \hline
   \multirow{3}{*}{Predictor} & \multirow{3}{*}{Timescale} & \multicolumn{6}{c|}{\textit{Avenue}} & \multicolumn{6}{c}{\textit{ShanghaiTech}} \\\cline{3-14}
   & & \multicolumn{3}{c|}{Summed Error $\mathcal{S}$} & \multicolumn{3}{c|}{Flattened Error $\mathcal{F}$} & \multicolumn{3}{c|}{Summed Error $\mathcal{S}$} & \multicolumn{3}{c}{Flattened Error $\mathcal{F}$}\\
    & & $m_1$ & $m_2$ & $m_3$ & $m_1$ & $m_2$ & $m_3$ & $m_1$ & $m_2$ & $m_3$ & $m_1$ & $m_2$ & $m_3$ \\
 \hline
 \multirow{4}{*}{LSTM} & $\tau=\delta=3$	    &{\textcolor{blue}{0.631}} &0.617 &0.586 &\textcolor{blue}{0.621} &0.606 &0.586 &0.610 &0.614 &0.550 &0.611 &0.616 &0.551 \\
& $\tau=\delta=5$	   &0.500 &0.414 &0.588 &0.500 &0.418 &0.588 &0.622 &0.623 &0.546 &0.626 &0.629 &0.550 \\
& $\tau=\delta=13$	    &0.567 &0.569 &0.593 &0.562 &0.568 &0.596 &0.624 &\textcolor{blue}{0.633} &0.534 &0.646 &\textcolor{blue}{0.655} &0.540 \\
& $\tau=\delta=25$	   &0.547 &0.551 &0.609 &0.526 &0.534 &0.605 &0.613 &0.618 &0.515 &0.637 &0.644 &0.534 \\
 \hline
\multirow{4}{*}{BiTraP (proposed)}& $\tau=\delta=3$ &0.522 &0.523 &0.643 &0.528 &0.528 &0.647 &0.659 &0.660 &0.632 &0.670 &0.671 &0.639 \\
& $\tau=\delta=5$ &0.543 &0.545 & {\textcolor{red}{0.718}} &0.550 &0.551 &\textbf{\textcolor{red}{0.720}} &0.674 &0.675 &0.642 &0.687 &0.688 &0.652 \\
& $\tau=\delta=13$	&0.554 &0.556 &0.690 &0.553 &0.555 &0.695 &0.689 &{\textcolor{red}{0.691}} &0.652 &0.716 &\textbf{\textcolor{red}{0.719}} &0.675 \\
& $\tau=\delta=25$	&0.548 &0.549 &0.684 &0.561 &0.564 &0.690 &0.664 &0.668 &0.623 &0.686 &0.690 &0.647 \\ \hline
\end{tabular}
\end{table*}

\begin{figure*}[h!]
  \centering
  \includegraphics[width=0.9\textwidth,trim=0 0 0 0,clip]{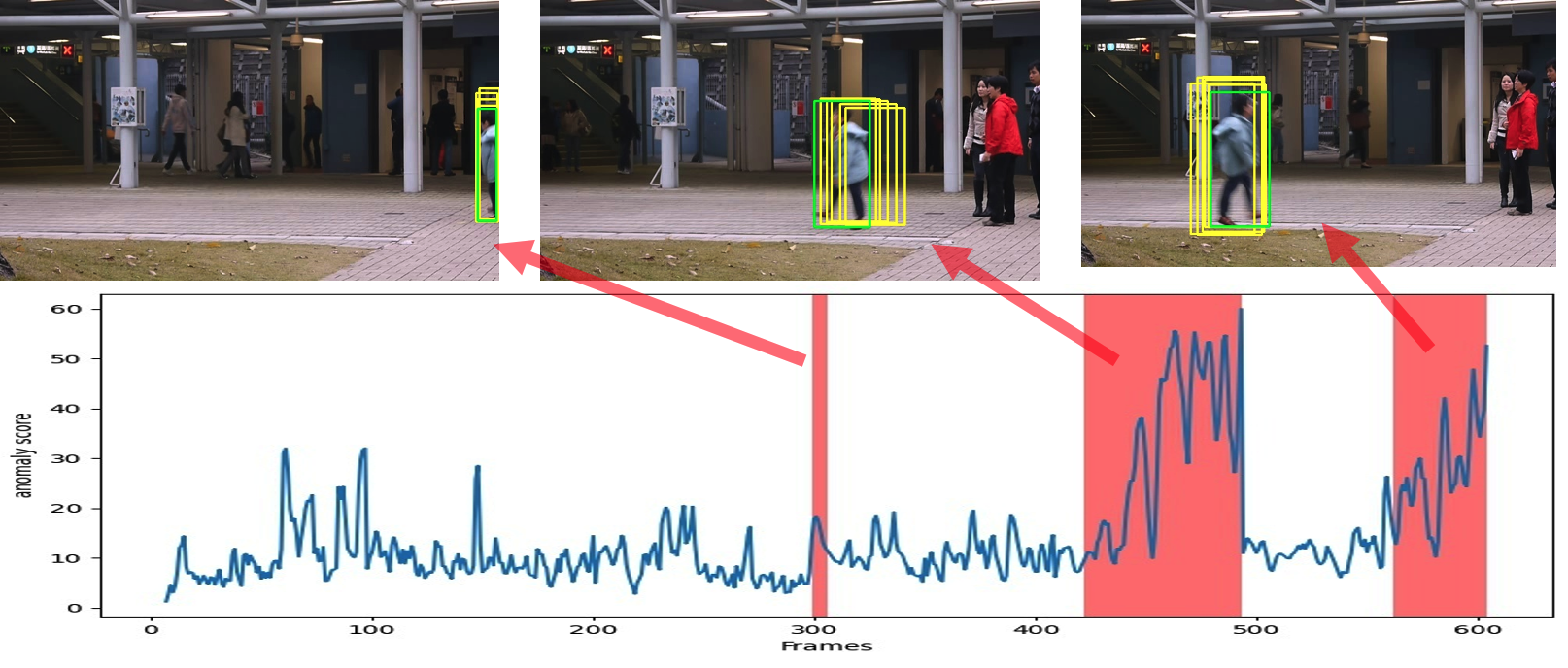}   
  \caption{A plot showing detected anomaly scores across all frames in a video and corresponding pedestrian activities in the image. The green bounding box is the ground truth and yellow bounding boxes are predicted trajectory (bounding box) sequences ($\tau=\delta=5$). The red shaded regions correspond to the pedestrian activities shown in the images above.} 
  \label{fig:anomaly_score_frames}
\end{figure*}

\subsection{Results and Discussions}
To verify the effectiveness of using BiTraP as a trajectory predictor and how the predictor affects the detection performance, we substitute the predictor with a seven-layer stacked long short-term memory (LSTM) network \cite{hochreiter1997long} while keeping the localization and detection modules the same as in our proposed pipeline. One may also think of this experiment as an ablation on the predictor module to test the effectiveness of our proposed BiTraP predictor in our pipeline. For the comparison stacked LSTM method, the first five LSTM layers contain 4, 3, 6, 4, and 4 units with sequences returned. The sixth LSTM layer consists of 4 units without returned sequences. The last layer was a fully connected layer with an adjustable output based on predicted trajectory sequence length $\delta$. A learning rate of 8.726e-06 was used for the LSTM model. We varied the number of layers, units and learning rate and obtained the above parameters giving the best detection results empirically. 

For both our pipeline using BiTraP as the predictor and the LSTM comparison model, we conducted experiments at different timescales $\tau=\delta=\{3, 5, 13, 25\}$. These timescales are set as suggested in \cite{rodrigues2020multi} so that  events that last varying sequence lengths can be observed and captured. For example, an activity such as jumping may last only a few timesteps (frames) while loitering is a longer-term anomaly. We report ROC and AUC detection results across these varying timescales. 

Table~\ref{table:frame_auc_error_all} presents the frame-level ROC AUC results comparing our pipeline using BiTraP trajectory predictor versus LSTM predictor, for both summed error and flattened error with all three types of error measures as described in Section~\ref{sec:predictor}. We can draw several observations from these results. First, as shown, our proposed method with BiTraP predictor outperforms LSTM by a large margin (6-10\%) and obtains the highest ROC AUC detection performance ($0.720$ for \textit{Avenue} and $0.719$ for \textit{ShanghaiTech}). Second, for both datasets tested, the ``flattened error'' method of handling sub-sequence prediction outperforms the ``summed error'' method, indicating that our approach of combining multiple predictions across overlapping sub-sequence predictions by sliding window method is effective and can aggregate prediction error information across multiple sub-sequence predictions to improve the anomaly detection performance.  Third, among the three types of error measurements (1-IOU, 1-GIOU and L2 norm), L2-norm seems to pair well with BiTraP predictor on the \textit{Avenue} dataset and 1-GIOU works well on the \textit{ShanghaiTech} dataset. IOU-based measures ($m_1$ and $m_2$) seem to work well with LSTM results, while L2 norm and GIOU-based measures ($m_2$ and $m_3$) work the best with BiTraP predictor. Additionally, when comparing across varying timescales,  we observe that longer timescales generally achieves higher detection performance, particularly for BiTraP ($\tau=\delta=5$ is best for \textit{Avenue} and $\tau=\delta=13$ for \textit{ShanghaiTech}, largely outperforming $\tau=\delta=3$ for example). One explanation can be that the \textit{ShanghaiTech} datasets include more long-term activities such as biking and loitering while \textit{Avenue} include, in comparison, shorter anomalous activities such as jumping or throwing. When $\tau$ and $\delta$ increase to 25, however, the AUC results dropped slightly, possibly due to the smaller number of sequences available with such long observed and prediction horizons (e.g., 33,472 sequences for when $\delta=25$ versus 82,230 sequences for when $\delta=3$ in \textit{Avenue} dataset).

As mentioned in section II-C, our experiments shown in table \ref{table:frame_auc_error_all} have a sliding window size of one. To further support our observation that combining multiple overlapping sub-sequences can help aggregate prediction error and improve anomaly detection, we performed an ablation study where we changed the sliding window size to 3, 5, 13, and 25 for timescales $\tau=\delta=3, 5, 13, 25$ respectively. Our obtained results in the ablation study show that the best frame-level AUC for both datasets decreases from 0.720 to 0.676 (BiTraP predictor, $\tau=\delta=13$, ``Flattened Error", L2 norm )  for \textit{Avenue} and 0.719 to 0.662 (BiTraP predictor, $\tau=\delta=13$, ``Flattened Error", 1-GIOU)  for \textit{ShanghaiTech}. We infer that the frame-level AUC decrease is a result of 1) eliminated sub-sequence overlap and 2) decreased trajectory sequences associated with a larger sliding window.



Figure~\ref{fig:anomaly_score_frames} shows an example plot of our detected anomaly scores across all frames in a video (video \#7, \textit{Avenue} dataset, $\tau=\delta=5$, flattened error with $m_3$ L2-norm-based error measure). The red highlighted segments in the plot represent the anomalous frames in the video. The top of the figure shows three example images that correspond to the three segments. In the first example, the pedestrian just came into view of the frame and is walking, thus our predicted trajectory bounding boxes (yellow) and the ground truth (green) align quite well and the anomaly score is relatively low. As time moves forward, the pedestrian started jumping/skipping/facing the other direction  (labeled as anomalous activities) instead of continuing the regular walking motion, thus, the predictions deviate further from the ground truth and the anomaly scores for the second and third examples went up higher to indicate the occurrence of the anomalous event.

We also provide a comparison of our frame-level detection performance with existing methods in the literature. As shown in Table~\ref{table:auc_compare}, our proposed method with the BiTraP trajectory predictor outperforms using stacked LSTM as predictor as well as both previous reconstruction-based approaches, the Conv-AE \cite{hasan2016learning} and TSC sRNN \cite{luo2017revisit}. However, the simple bounding box coordinate-based representation ($4\times1$) used in the current pipeline does not seem to be comprehensive enough compared with the more complicated features used in alternative prediction-based works, such as the intensity and gradient difference for all frame pixels in the U-Net \cite{liu2018future}, the full-body skeleton representations as used in MPED-RNN \cite{Morais_2019_CVPR}, or the fusion of pose features ($50\times1$) across multiple timescales \cite{rodrigues2020multi}. Nevertheless, our work shows that it is possible to perform anomaly detection using only a very simple and efficient bounding box trajectory representation and that our unsupervised, prediction-error-based pipeline outperforms prior reconstruction-based methods. Future work will include incorporating additional motion, pose, and appearance/texture features from the pedestrians as well as the environmental contexts to further improve the detection performance.

\begin{table}[h!]
\centering
\caption{AUC comparison results with existing methods.}
\label{table:auc_compare}
\begin{tabular}{c|c|c} 
 \hline
  & \textit{Avenue} & \textit{ShanghaiTech} \\ 
 \hline
Stacked LSTM as Predictor (baseline) & 0.631 & 0.655\\
Conv-AE \cite{hasan2016learning} & 0.702 & 0.704  \\ 
TSC sRNN \cite{luo2017revisit} & 0.817 & 0.680\\
 \textbf{Ours with BiTraP Trajectory Predictor} & 0.720 & 0.719 \\
\rowcolor{gray!30} U-Net Frame Prediction \cite{liu2018future}  & 0.849 & 0.728  \\
\rowcolor{gray!30} MPED-RNN \cite{Morais_2019_CVPR}& 0.863 & 0.734  \\
\rowcolor{gray!30} Multi-Timescale Pose Prediction \cite{rodrigues2020multi} & 0.828 & 0.760  \\ 
 \hline
\end{tabular}
\end{table}


\subsection{Computation Time}
All experiments were conducted on a desktop computer with Intel Xeon CPU E5-2698 v4 @ 2.20GHz and a single Telsa V100-SXM2 16 GB GPU. On the \textit{Avenue} testing dataset, our implementation runs at $\sim$7.3 FPS (frames per second) with $\tau=\delta=3$ and $\sim$5.5 FPS with $\tau=\delta=25$. Within the pipeline, the localization step by deep SORT takes approximately 76\% of the total running time, the BiTraP trajectory prediction takes approximately 22\% of the total time (in which 1.6\% is spent on loading the trajectory data, 16.4\% for inference and  4.0\% for saving the models), and the computation of error measures and anomaly scores takes approximately 2\%. Note that the localization step can be done once for all frames at the beginning of the testing pipeline off-line, and has potential for further speed-up for on-line applications (the original SORT method \cite{bewley2016simple} reported 60Hz runtime and deep SORT at 40Hz \cite{wojke2017simple}; in our experiments we obtained approximately 10.5Hz). The BiTraP inference and anomaly score calculation are both computationally efficient and can operate at real time.

\section{Conclusion}
In this work, we presented a trajectory localization and prediction based pipeline for unsupervised pedestrian video anomaly detection. Our bounding box-based trajectory representation provided a simple, efficient, and interpretable approach to localize and detect anomalous activities in pedestrians. We show that our prediction-based pipeline can outperform previous reconstruction-based methods. Our proposed detection pipeline does not require ground truth anomaly labels for training and depends on the video data inputs alone. We also verified the effectiveness of our proposed BiTraP predictor in the pipeline on varying timescales. 

Current datasets regard pedestrian walking sequences as ``normal'' and tag other non-walking activities, such as jumping, running, and biking as anomalies. However, depending on the environmental context, further analysis can be conducted to recognize the semantics of pedestrian activities in more detail. For example, a child jumping on the playground may be perfectly ``normal'' but a child jumping in front of a car should be flagged and cautioned as anomalous. Additional motion, pose, and appearance/texture features from the pedestrians as well as the environmental contexts can be incorporated to reflect such semantic meaning and improve the detection performance. Further refinement on localizing the anomalous events (e.g., on a pixel-level) and extension of the current framework to non-stationary camera data (e.g., as collected from an autonomous vehicle) can also be explored.




\section*{Acknowledgment}
This work was supported by a grant from Ford Motor Company via the Ford-UM Alliance under award N028603. This material is based upon work supported by the Federal Highway Administration under contract number 693JJ319000009. Any options, findings, and conclusions or recommendations expressed in the this publication are those of the author(s) and do not necessarily reflect the views of the Federal Highway Administration.


\bibliographystyle{IEEEtran}
\bibliography{reference.bib}

\end{document}